\documentclass[10pt,times,mathptm,psfig,twocolumn,journals]{IEEEtran}
\usepackage{url}
\usepackage{amssymb}
\usepackage{array}
\usepackage{amsmath}
\usepackage{graphicx}
\usepackage{booktabs}
\usepackage{epsfig}
\usepackage{ulem}
\usepackage{hyperref}
\hypersetup{pdfborder={0 0 0}, colorlinks=true, linkcolor=red, citecolor=blue}
\usepackage{subfigure}
\usepackage{cite}
\usepackage{epstopdf}
\usepackage{amsmath}
\usepackage[misc]{ifsym}
\usepackage{bm}
\usepackage{mathrsfs}
\usepackage{float}
\usepackage{setspace}
\usepackage{multirow}
\usepackage{tabularx}
\usepackage{url}
\usepackage{booktabs}
\usepackage{makecell}
\usepackage{epigraph}
\usepackage{bbding}
\usepackage{txfonts}
\usepackage{gensymb}

\begin{document}

\title{Deep Rotation Correction without Angle Prior}

\author{Lang~Nie, Chunyu~Lin,~\IEEEmembership{Member,~IEEE}, Kang~Liao, Shuaicheng Liu,~\IEEEmembership{Member,~IEEE}, \\Yao~Zhao,~\IEEEmembership{Fellow,~IEEE}
\thanks{
\textit{Corresponding author: Chunyu Lin}}
\thanks{Lang Nie, Chunyu Lin, Kang Liao, Yao Zhao are with the Institute of Information Science, Beijing Jiaotong University, Beijing 100044, China, and also with the Beijing Key Laboratory of Advanced Information Science and Network Technology, Beijing 100044, China (email: nielang@bjtu.edu.cn, cylin@bjtu.edu.cn, kang\_liao@bjtu.edu.cn, yzhao@bjtu.edu.cn).}
\thanks{Shuaicheng Liu is with School of Information and Communication Engineering, University of Electronic Science and Technology of China, Chengdu, 611731, China (liushuaicheng@uestc.edu.cn).}
}


\maketitle
\begin{abstract}
    Not everybody can be equipped with professional photography skills and sufficient shooting time, and there can be some tilts in the captured images occasionally.
    In this paper, we propose a new and practical task, named \underline{Rotation Correction}, to automatically correct the tilt with high content fidelity in the condition that the rotated angle is unknown. This task can be easily integrated into image editing applications, allowing users to correct the rotated images without any manual operations.
    To this end, we leverage a neural network to predict the optical flows that can warp the tilted images to be perceptually horizontal. Nevertheless, the pixel-wise optical flow estimation from a single image is severely unstable, especially in large-angle tilted images. To enhance its robustness, we propose a simple but effective prediction strategy to form a robust elastic warp. Particularly, we first regress the mesh deformation that can be transformed into robust initial optical flows. Then we estimate residual optical flows to facilitate our network the flexibility of pixel-wise deformation, further correcting the details of the tilted images.
    To establish an evaluation benchmark and train the learning framework, a comprehensive rotation correction dataset is presented with a large diversity in scenes and rotated angles. Extensive experiments demonstrate that even in the absence of the angle prior, our algorithm can outperform other state-of-the-art solutions requiring this prior. The code and dataset are available at \url{https://github.com/nie-lang/RotationCorrection}.
\end{abstract}

\markboth{}
{Shell \MakeLowercase{\textit{et al.}}: Bare Demo of IEEEtran.cls for IEEE Transactions on Magnetics Journals}
\IEEEpeerreviewmaketitle

\begin{figure*}[!t]
    \centering
    \includegraphics[width=0.97\textwidth]{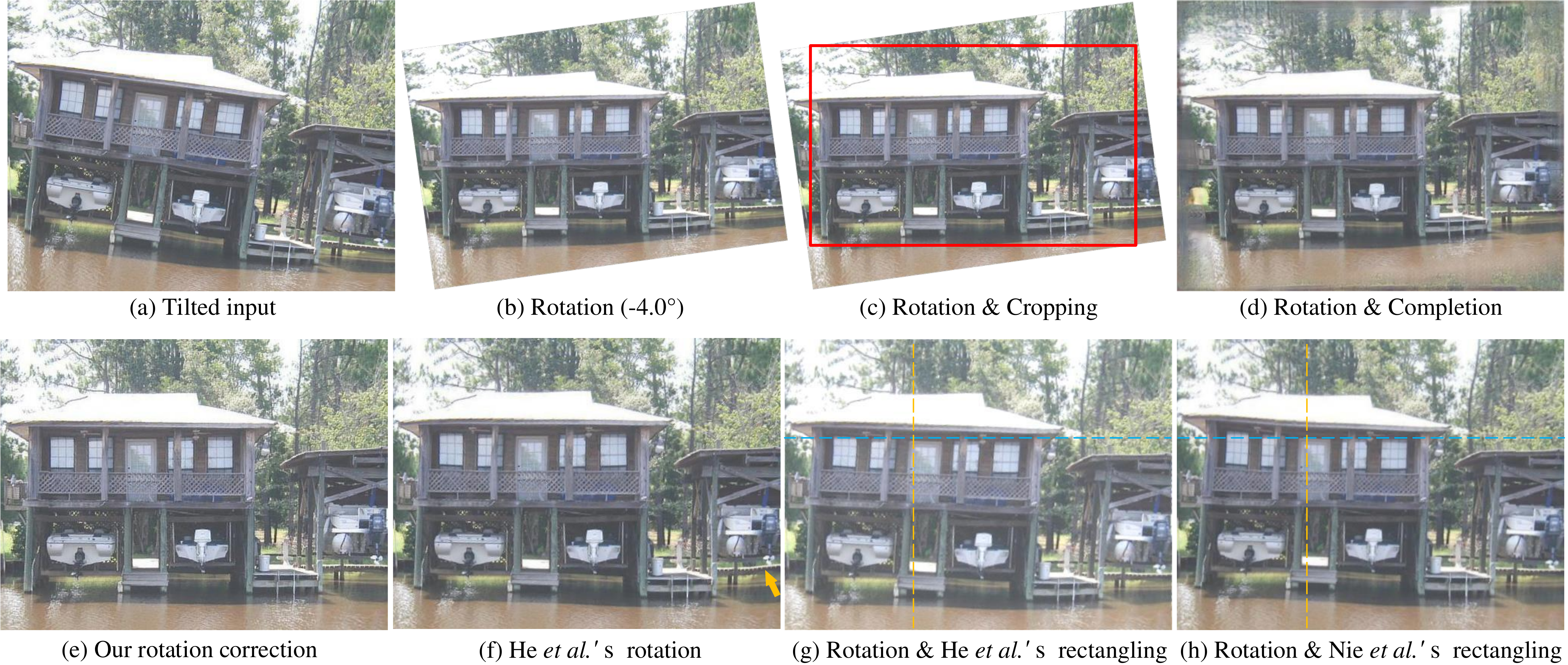}
    \caption{Different solutions to correct the tilted image. Our solution (e) can eliminate the tilt without angle prior, while the others (b)(c)(d)(f)(g)(h) require an accurate rotated angle. The red square denotes the cropping region, and the arrow highlights the distorted area. The horizontal and vertical dotted lines are drawn to help observe the slight tilt.}
    \label{fig1}
\end{figure*}

\section{Introduction}
\label{intro}
People favor recording fascinating landscapes and objects by taking photos. To obtain a visually pleasing appearance, they have to adjust the shooting perspective carefully, and firmly hold the camera to catch a perceptually horizontal photograph. However, limited by the photography skills and shooting time, the captured images might exhibit some tilts (Fig.\ref{fig1}{\color{red}a}) sometimes. To overcome the annoying tilt, people have to rotate the images by manually fine-tuning the rotated angle. Nevertheless, this rotation operation is rigid, which destroys the rectangular boundaries (Fig.\ref{fig1}{\color{red}b}).

To obtain the visually horizontal perception and retain the rectangular boundaries simultaneously, cropping and completion \cite{yi2020contextual} are the common operations following the rigid rotation. As illustrated in Fig. \ref{fig1}{\color{red}c} and Fig. \ref{fig1}{\color{red}d}, these operations decrease or increase the image contents,  damaging the authenticity of images.

To avoid this problem, image rectangling algorithms \cite{he2013rectangling, li2015geodesic, zhang2020content, nie2022deep, wu2022rectangling} can be leveraged to warp the rotated image (Fig.\ref{fig1}{\color{red}b}) to rectangular image. Fig. \ref{fig1}{\color{red}g} and Fig. \ref{fig1}{\color{red}h} demonstrate the results of He $et\ al.$'s traditional rectangling \cite{he2013rectangling} and Nie $et\ al.$'s deep rectangling \cite{nie2022deep}. But these solutions might reintroduce some slight tilts into the rectangling results because they neglect the angle-preserving constraint. Besides that, content-aware rotation \cite{he2013content} is another practical solution. Given the prior angle, it directly warps the contents in a tilted image to produce a visually horizontal perception without content increasing or decreasing. But the warp optimization heavily relies on the performance of the line segment detector (LSD\cite{von2008lsd}), inevitably yielding local distortions where LSD fails (Fig. \ref{fig1}{\color{red}f}). Moreover, both rectangling-based solutions and the content-aware rotation scheme share a common limitation: the rotated angle should be known in advance. This limitation makes them non-automatic and multi-step solutions.

In this paper, we propose a new and practical task, Rotation Correction, aiming to solve the above problems in one step. To be rigorous, we define this task as automatically correcting the \uline{2D in-plane tilt} (roll) with \uline{high content fidelity} (preserving contents and boundaries) without the angle prior. It can be easily integrated into image edition-related applications, freeing the users to rectify the tilts from any manual operations.

To achieve this goal, we design a simple but effective neural network to predict the optical flows progressively that can warp the tilted image to be perceptually horizontal. Nevertheless, the pixel-wise prediction in large-angle tilted scenes is extremely not robust, which requires the flows to be large and stable. To enhance the robustness, we first propose to predict the mesh deformation so that every pixel in a grid corresponds to an identical homography transformation \cite{9605632}. The estimated mesh deformation can be more robust and stable due to its lower resolution representation than the optical flow. Subsequently, we transform the mesh deformation into the robust initial optical flows and predict the residual flows to facilitate our network the flexibility of pixel-wise deformation. In this progressive manner, our predicted flows are both robust and elastic, yielding better correction in detail.

Actually, predicting such flows from a single image is an ill-pose problem, which requires a high-quality dataset to assist in learning this prediction capability. Meanwhile, to establish an evaluation benchmark, we build a comprehensive rotation correction dataset (DRC-D) with a large diversity in scenes and tilted angles. Particularly, we leverage He $et\ al.$'s rotation \cite{he2013content} to generate abundant sample candidates, and further filter and rectify them manually. In sum, the proposed dataset includes over 6$k$ samples with tilted inputs, corrected labels, and rotated angles.

Experimental results show that our ``no-angle prior" algorithm outperforms the existing state-of-the-art solutions requiring this prior, quantitatively and qualitatively. We conclude our contributions as follows:
\begin{itemize}
    \item We propose a new and practical rotation correction task, aiming to automatically correct the 2D in-plane tilt with high content fidelity without the angle prior. To accomplish it, an automatic solution is proposed to rectify the tilts by predicting the optical flows for warping.
    \item 
    To address the instability of monocular optical flow prediction, we propose a simple but effective strategy to form a robust elastic warp, by integrating the robustness of sparse mesh estimation into dense optical flow prediction.
    \item Due to the absence of a proper dataset with tilted and corrected images, we build a rotation correction dataset with a wide range of rotated angles and scenes.
 \end{itemize}

The remainder of this paper is organized as follows. In Section \ref{related_work}, we discuss the related works of rotation correction. The methodology and dataset are described in Section \ref{method} and \ref{dataset}, respectively.
In Section \ref{experiment}, we demonstrate extensive experiments to validate the effectiveness of the proposed solution. Finally, we conclude this work in Section \ref{conclusion}.

\section{Related Work}
\label{related_work}
We review the related methods on image rotation, rectangling, and monocular optical flow estimation here.
\subsection{Image Rotation}
\label{rotation}
The image rotation process can be implemented in two steps: estimating a rotated angle and rotating.

For the first step, people can acquire the rotated angle by rotating an image clockwise or counterclockwise until it becomes perceptually horizontal. But this manner demands the manual operation and time cost, especially for the images with few horizontal/vertical straight lines.
An alternative way is to estimate the rotated angle from a single image. Particularly, some methods detect the vanishing points  \cite{gallagher2005using, lee2021ctrl, lin2022deep} first and determine the horizon accordingly. Others devote themselves to directly calibrating the camera parameters \cite{xian2019uprightnet, lee2013automatic, do2020surface} such as camera poses that indicate the rotated angle.

Then, the rotating operation is conducted. In addition to the rigid rotation, other relatively complex transformations (e.g., homography) \cite{lee2013automatic, do2020surface} can be adopted to upright the tilted image. However, they inevitably destroy the rectangular boundaries.
To avoid it, He $et\ al.$ propose the content-aware rotation \cite{he2013content} to preserve the boundaries while exhibiting a perceptually rotated appearance. It leverages a property that the human eyes are sensitive to
tilted horizontal/vertical lines, optimizing a mesh wap that encourages the straight lines to rotate at the same angle.

In contrast, our purpose is to design a one-step solution to correct the tilt automatically with high content fidelity.

\subsection{Image Rectangling}
\label{rectangling}
Rectangling \cite{he2013rectangling, li2015geodesic, zhang2020content, nie2022deep, wu2022rectangling} refers to the problem of correcting the irregular boundaries (e.g. the rigid rotated images, panoramas, etc) to a rectangle. To this end, He $et\ al.$ \cite{he2013rectangling} propose a two-stage warping solution to acquire an initial mesh and optimize a line-preserving target mesh. However, it can only protect limited perceptual properties (such as straight lines), usually failing in scenes with abundant non-linear structures. Recently, Nie $et\ al.$ \cite{nie2022deep} combine this traditional warping problem with deep learning to facilitate the algorithm the perception capability to abundant semantic properties. It significantly reduces the distortions around non-linear objects while simplifying the rectangling process to a one-stage pipeline.

\subsection{Monocular Optical Flow}
\label{warping}
Monocular optical flow prediction can be widely used in extensive computer vision applications, such as wide-angle portrait correction \cite{tan2021practical}, rectangling \cite{nie2022deep}, retargeting \cite{tan2019cycle}, stitching\cite{kweon2021pixel}, etc.

Tan $et\ al.$ \cite{tan2021practical} predict the optical flows to correct the wide-angle portrait distortions. These predicted flows are robust and stable because wide-angle distortions are relatively slight, which does not require long optical flows. In image stitching, Kweon $et\ al.$ \cite{kweon2021pixel} propose to estimate the flows for non-overlapping regions. The prediction is severely unstable because it requires predicting long optical flows to satisfy the long-range warping.
In large-angle tilted images, long flows are also required. In this paper, we enhance the robustness of monocular optical flow prediction by integrating the robustness of sparse mesh estimation into dense optical flow prediction.

\section{Methodology}
\label{method}
In this section, we first discuss the robust elastic warp in Section \ref{method_1}. Then we propose our solution to rotation correction in detail in Section \ref{method_2} and Section \ref{method_3}. Finally, we clarify the difference between related works and ours in Section \ref{method_4}.

\subsection{Robust Elastic Warp}
\label{method_1}
\begin{figure}[!t]
    \centering
    \includegraphics[width=0.48\textwidth]{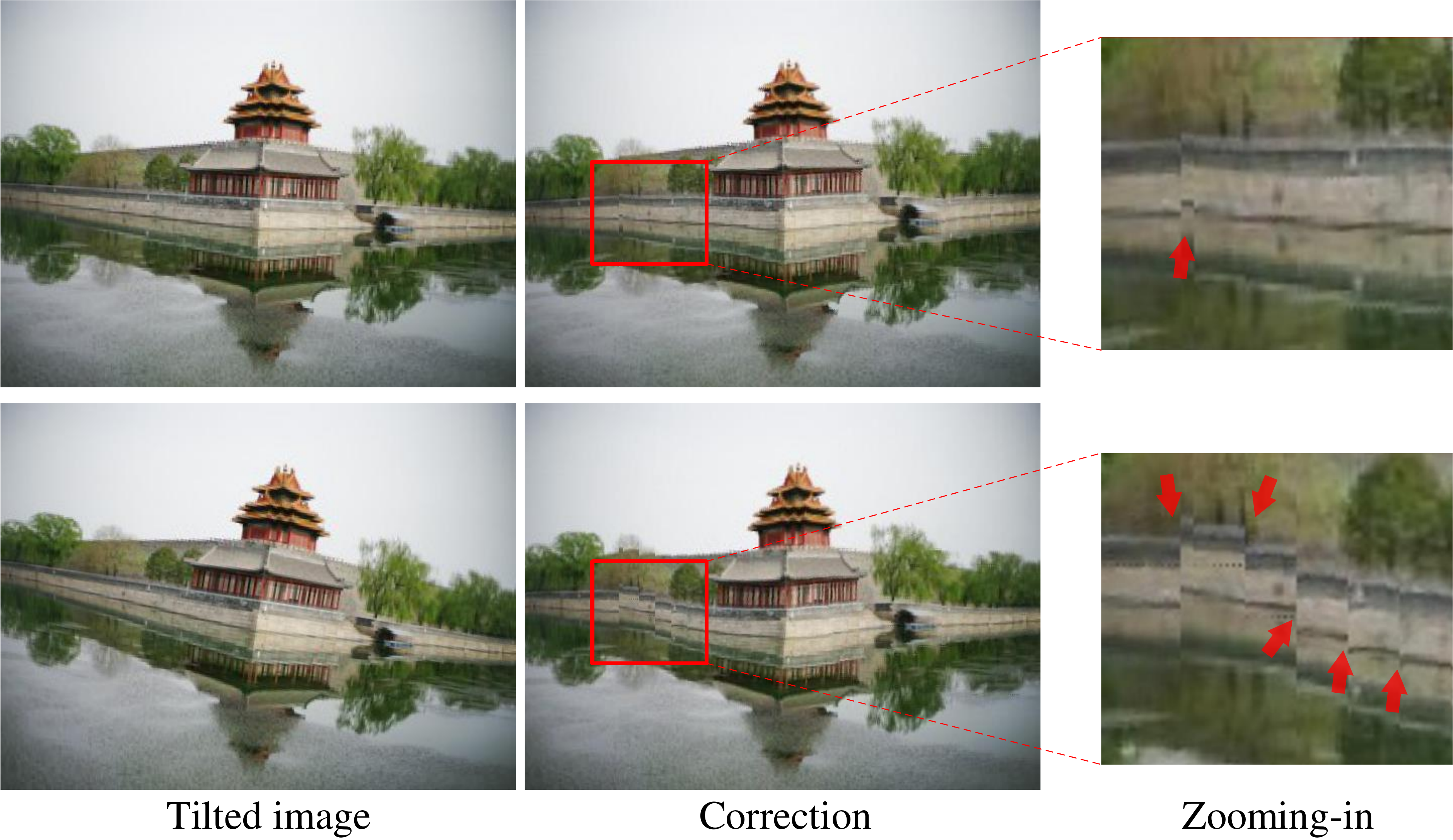}
    \vspace{-0.1cm}
    \caption{Qualitative analysis of the optical flow robustness. Top: small-angle tilt (3.1\degree). Bottom: large-angle tilt (9.5\degree).}
    \label{flow_robust_fig}
\end{figure}

Considering the flexibility of optical flows, we apply monocular optical flow estimation to realize rotation correction. At first, we adopt a simple Unet-style \cite{ronneberger2015u} network with 5 different resolution hierarchies to predict the optical flows in our proposed dataset. The results are shown in Fig. \ref{flow_robust_fig}, where two instances with different tilted angles are exhibited and the red arrows highlight the discontinuous regions. From the corrected results, we can observe:
\begin{itemize}
    \item The performance of monocular optical flow estimation is not stable. Even in a case of small-angle tilt, the discontinuous content could appear.
    \item As the increase of tilted angle, the predicted flows become increasingly unstable. More discontinuous regions counld be observed in the case of large-angle tilt.
 \end{itemize}
The cause of unstable performance can attribute to the high resolution of optical flows (the same resolution as the input image). It's challenging to predict accurate flows for every pixel, especially accurate long flows in large-angle tilted cases.

To enhance the robustness of optical flows, we propose a robust elastic warp from low-resolution robust mesh deformation to high-resolution flexible optical flows. Particularly, the network is designed to regress the mesh deformation first. Due to the low-resolution representation of mesh, the burden of network prediction is reduced from the flows of full pixels to the offsets of sparse mesh vertexes, thus ensuring the robustness of mesh deformation prediction.
When the mesh deformation is transformed into corresponding high-resolution optical flows, the converted flows are naturally robust.
But it sacrifices the flexibility of pixel-wise deformation. Subsequently, residual optical flows can be predicted to make up for this defect.

\subsection{Network Architecture}

\label{method_2}
\begin{figure*}[!t]
    \centering
    \includegraphics[width=0.97\textwidth]{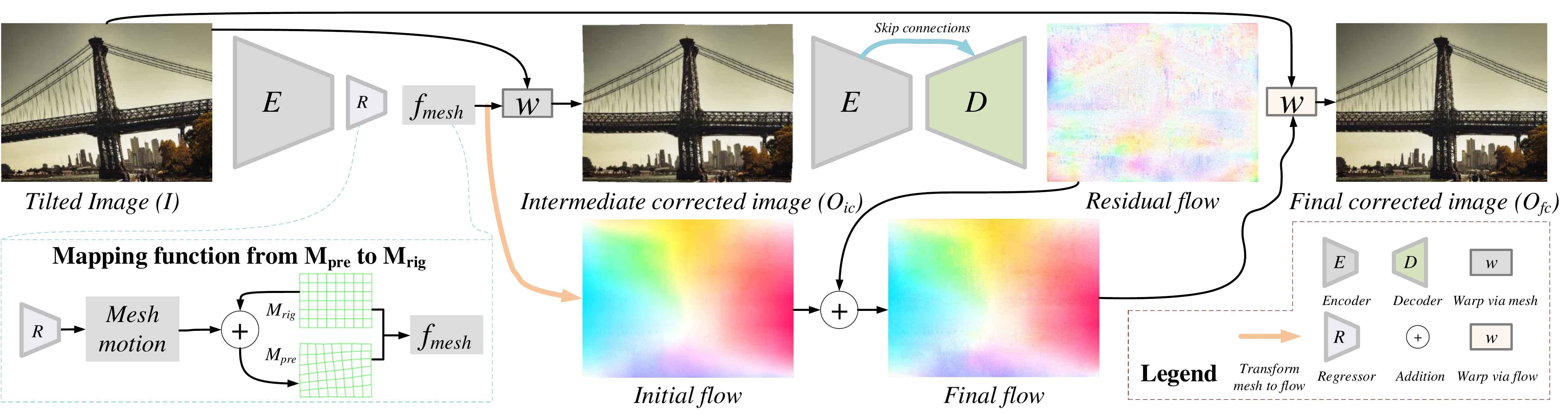}
    \vspace{-0.3cm}
    \caption{The pipeline of the proposed rotation correction network. We first regress the low-resolution mesh deformation that can be transformed into a robust initial flow. Then the residual flows are predicted to make up for the flexibility of pixel-wise deformation.}
    \label{network}
\end{figure*}

To embody the effectiveness of this strategy, we accomplish it with a simple network instead of designing complex network structures. The network pipeline is illustrated in Fig. \ref{network}.

\begin{figure}[!t]
    \centering
    \includegraphics[width=0.45\textwidth]{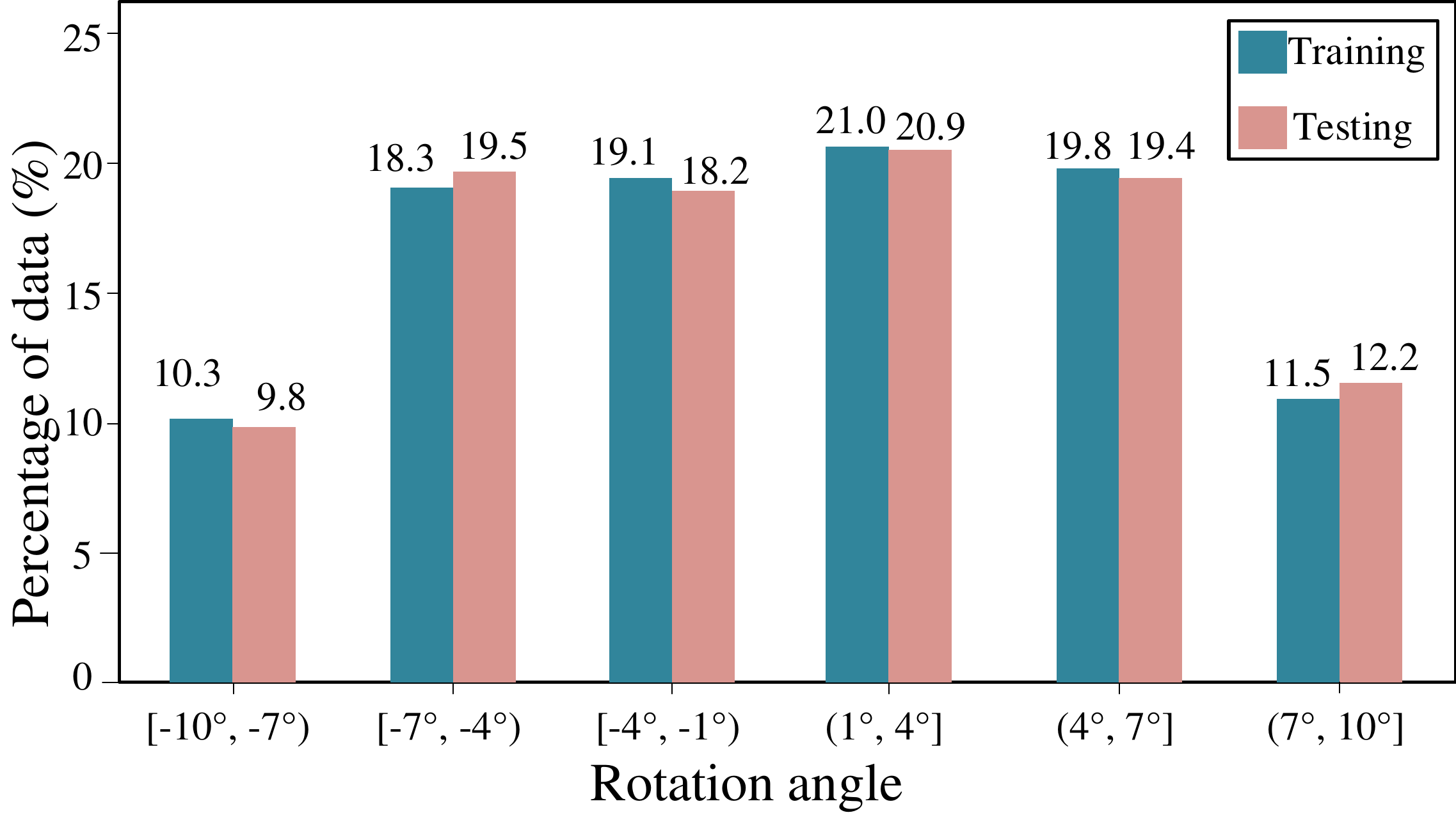}
    \vspace{-0.4cm}
    \caption{Data distribution of training and testing set of the proposed dataset with respect to rotated angles.}
    \label{dataset_distribution}
\end{figure}

\vspace{0.1cm}
\subsubsection{Mesh Prediction} Assuming the mesh resolution is $U\times V$, the mesh vertexes have a resolution of $(U+1)\times (V+1)$. We predefine a rigid mesh $M_{rig}$ and place it on the corrected image. Then our goal is to predict a mesh $M_{pre}$ that is placed on the tilted image. Supposing the position of a vertex in $M_{rig}$ is $(m_{ij}, n_{ij})$, we desisn an encoder and regressor to predict the corresponding vertex motion $(\Delta m_{ij}, \Delta n_{ij})$. The corresponding vertex position in $M_{pre}$ is denoted as $(m_{ij}+\Delta m_{ij}, n_{ij}+\Delta n_{ij})$.

For the encoder, we employ 10 convolutional layers with filter numbers set to 64, 64, 64, 64, 128, 128, 128, 128, 256, and 256. A max-pooling layer is adopted every two convolutions except at the beginning and the end. For the regressor, we stack 4 convolutional layers with max-pooling operations and 3 fully connected layers to regress the vertex motion. The filter numbers of convolutions and dimensions of fully connected layers are set to 256, 256, 512, 512, 2048, 1024, and $(U+1)\times (V+1)\times 2$, respectively.

\vspace{0.1cm}
\subsubsection{Mesh to Flow} We transform the mesh deformation into optical flows in this step.
Particularly, we calculate a homography transformation $H_{uv} (u=1,2,...,U; v=1,2,...,V)$ for every corresponding grid pair from $M_{rig}$ to $M_{pre}$ with respective four adjacent vertexes by direct linear transform (DLT) algorithm \cite{hartley2003multiple}. 
We represent $H_{uv}$ as a $3\times 3$ matrix as follows:
\begin{equation}
    H_{uv}=\begin{pmatrix}
        h_1 \\
        h_2\\
        h_3
        \end{pmatrix}=\begin{pmatrix}
            a_{11}&a_{12}& a_{13} \\
            a_{21}& a_{22}& a_{23}\\
            a_{31}& a_{32} & 1
    \end{pmatrix},
    \label{eq1}
 \end{equation}
where $h_1$, $h_2$ and $h_3$ are the row vectors of $H_{uv}$.
Then, for every pixel ($x,y$) in the corrected image at the $uv$-th grid of $M_{rig}$, we convert the mesh deformation into the corresponding flows ($f_{hor}^{xy},f_{ver}^{xy}$) as Eq. (\ref{eq2}):
\begin{equation}
    (f_{hor}^{xy},f_{ver}^{xy})=(\frac{h_1\cdot(x\ y\ 1)^T}{h_3\cdot(x\ y\ 1)^T}-x,\ \frac{h_2\cdot(x\ y\ 1)^T}{h_3\cdot(x\ y\ 1)^T}-y).
    \label{eq2}
 \end{equation}

 The initial flows calculated from the mesh deformation are stable because the mesh deformation is robust due to the low-resolution characteristic. However, the flows sacrifice the flexibility of pixel-wise deformation. For example, the flows transformed from the same grid pair share the same warping function (homography transformation).

\vspace{0.1cm}
\subsubsection{Residual Flow Prediction}
To remedy this drawback, we design an additional encoder-decoder network to predict the residual flows. This part takes the intermediate corrected image $O_{ic}$ as the input that can be obtained by warping the tilted image $I$ via the mesh deformation. We formulate this process as Eq. (\ref{eq3}):
\begin{equation}
    O_{ic} = \mathcal{W}(\{H_{uv}|u=1,2,...,U;v=1,2,...,V\}, I),
    \label{eq3}
 \end{equation}
where $\mathcal{W}(\cdot ,\cdot )$ represents the mesh warping operation that takes a mapping function and an image as input.

The encoder-decoder network follows a similar structure to Unet \cite{ronneberger2015u}. Specifically, its encoder shares the same structure as that in the mesh prediction stage. The decoder adopts a symmetrical structure as the encoder, where the transposed convolutions are employed to increase the feature resolution, and skip connections are used to connect the features with the same resolution. At the end of the decoder, an additional convolution with 2 filters is adopted to output the residual optical flows $(\Delta f_{hor}^{xy}, \Delta f_{ver}^{xy})$. By the addition of initial flows and residual flows, we obtain the final flows that are used to get the final corrected image $O_{fc}$ as Eq. (\ref{eq4}):
\begin{equation}
    O_{fc}(x,y) = I(x+f_{hor}^{xy}+\Delta f_{hor}^{xy}, y+f_{ver}^{xy}+\Delta f_{ver}^{xy}).
    \label{eq4}
 \end{equation}

\subsection{Objective Function}
\label{method_3}
The objective function consists of a content term $\mathcal{L}_{content}$ and a symmetry-equivariant term $\mathcal{L}_{symmetry}$, which is formulated as follows:
\begin{equation}
    \mathcal{L} = \mathcal{L}_{content} + \omega \mathcal{L}_{symmetry},
 \end{equation}
 where $\omega$ is the weight to balance the significance of $\mathcal{L}_{content}$ and $\mathcal{L}_{symmetry}$.

 \begin{spacing}{1.5}
 \end{spacing}
 \textbf{Content Term.}
We design our content term following two principles:

1) Simple. We hope the proposed robust elastic warp to be simple but effective. Therefore, the learning framework should be effectively trained by a simple loss.

2) Perceptual. The network should focus on the semantic properties that could embody the horizon instead of the ordinary pixels without attention.

To satisfy the above requirements, we leverage the perceptual loss \cite{johnson2016perceptual} as our loss function.
The perceptual loss minimizes the distance between the high-level semantics of corrected images and that of labels. It encourages the objects that are more significant in semantics instead of all pixels to be more strictly horizontal, which conforms to human perception naturally.
It can promote tilt correction from two complementary perspectives: 1) Find the semantically significant region by the pretrained VGG19 \cite{simonyan2014very}. Then, the network is encouraged to preserve the shape of these regions and only correct the tilt. 2) Find the semantically insignificant regions implicitly. Compared with semantically significant regions, these insignificant regions in perceptual loss have quietly low errors. Therefore, to keep a rectangular boundary, many warping operations such as stretching or flattening usually happen in these regions (e.g., the lake, sky, and so on), making the distortions visually unnoticeable.

We define $\varPhi(\cdot )$ as the operation of extracting the semantic features. It takes an image as input and outputs the feature maps from VGG19 \cite{simonyan2014very}. In our implementation, we use the features extracted after the $conv4\_3$ layer as an effective perceptual representation.
Denoted the corrected label as $\hat{O}$, the conent term can be formulated as follows:

\begin{equation}
    \mathcal{L}_{content} = \frac{1}{N} (\left \| \varPhi(\hat{O})-\varPhi(O_{ic}) \right \|_2^2 + \lambda\left \| \varPhi(\hat{O})-\varPhi(O_{fc}) \right \|_2^2),
 \end{equation}
 where $N$ and $\lambda$ denotes the element number of the feature maps and the weight for residual flow prediction, respectively.

 \begin{spacing}{1.5}
 \end{spacing}
 \textbf{Symmetry-Equivariant Term.}
For two left-right symmetrical images, the corresponding corrected images should also be symmetrical. In other words, if we exchange the operation orders of symmetry and rotation correction, the results would be invariant.
 Based on this observation, we design a symmetry-equivariant loss to further facilitate the network the capability of horizon perception. Assuming $I^{sym}$ be the left-right symmetric image of $I$, the corresponding outputs are $O_{ic}^{sym}$ and $O_{fc}^{sym}$. Then the symmetry-equivariant term can be defined as:
 \begin{equation}
    \begin{aligned}
        \mathcal{L}_{symmetry} &= \frac{1}{N} (\left \| \varPhi(O_{ic}^{sym})-\varPhi(f_{sym}(O_{ic})) \right \|_2^2 + \\\\
    &\lambda\left \| \varPhi(O_{fc}^{sym})-\varPhi(f_{sym}(O_{fc})) \right \|_2^2),
    \end{aligned}
 \end{equation}
where $f_{sym}(\cdot)$ denotes the operation of left-right symmetry.

\subsection{Comparisons to Related Work}
\label{method_4}
The proposed rotation correction method takes a tilted image as input. It rectifies the content tilt automatically without an angle prior requirement. Next, we discuss the differences between the related works and ours.

\textbf{Nie $et\ al.$'s rectangling\cite{nie2022deep} vs. ours:} 
The input of Nie $et\ al.$'s rectangling is the rigidly rotated image with a specific angle (e.g., the tilted angle). Then it is warped to produce rectangular boundaries using the mesh deformation.
However, the mesh representation loses the flexibility of pixel-wise deformation. In our work, we use optical flow to produce a flexible warp. To overcome the instability of monocular optical flow estimation especially in heavily tilted images, we propose to incorporate the robustness of mesh deformation with the flexibility of optical flows. This ingenious combination of their natural advantages yields a robust elastic warp.

\textbf{He $et\ al.$'s rotation\cite{he2013content} vs. ours:} The angle prior is also required in He $et\ al.$'s rotation, which takes a tilted image and the corresponding tilted angle as input. Then it optimizes an energy function that encourage the boundaries to keep rectangular and the lines to rotate with the known tilted angle. Different from it, our solution frees from the limitation of the angle prior, predicting a warp according to the different tilted degrees.

 \begin{figure*}[!t]
    \centering
    \includegraphics[width=0.96\textwidth]{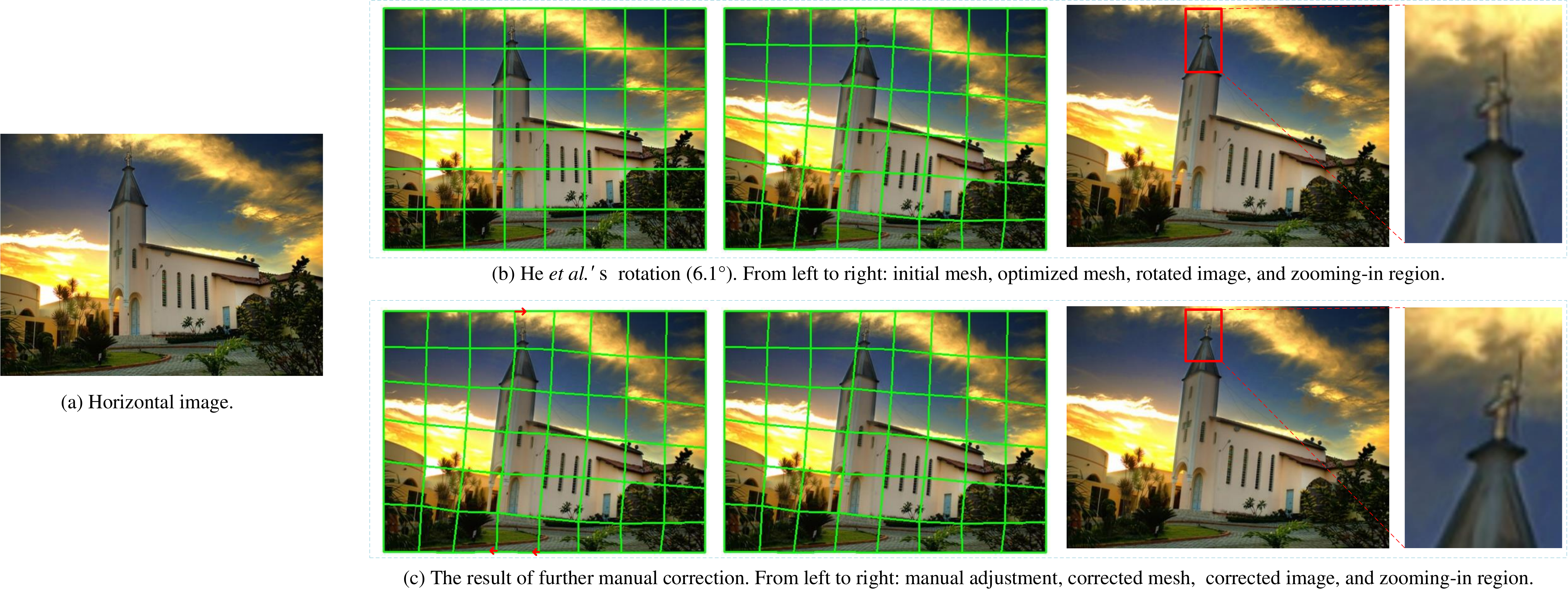}
    \caption{The process of dataset generation. We further correct the randomly rotated result generated from He $et\ al.$' rotation \cite{he2013content}. The red arrows in (c) indicate the manual adjustment of moving the mesh vertices. He $et\ al.$'s rotation neglects the rotation of the corss ((b) right), while our manual correction slightly rotates it to produce a more natural appearance ((c) right).}
    \label{dataset_generation}
\end{figure*}

\section{Data Preparation}
\label{dataset}
As there is no proper dataset for rotation correction, we build a comprehensive dataset to train the learning framework and establish an evaluation benchmark. The process can be divided into 4 steps:

1) We collect abundant perceptually horizontal images (Fig. \ref{dataset_generation}{\color{red}a}) as the ground truth of the rotation corrected images. Specifically, we select the categories that belong to buildings and landscapes from ImageNet \cite{deng2009imagenet}, e.g. boathouse, castle, church, lakeshore, volcano, etc, because the horizontal properties of these images are relatively easy to perceive. Then we manually pick up the perceptually horizontal images from these labeled candidates. We repeat this filtering process for 3 epochs and less than 3$k$ images remain from more than 10$k$ images. Besides, to enrich the variety of scenes, we also collected some pictures of other categories by ourselves.

2) We apply He $et\ al.$'s rotation algorithm \cite{he2013content} to rotate these horizontal images (Fig. \ref{dataset_generation}{\color{red}b}). Particularly, every image is rotated with 6 random angles that belong to 6 different angle intervals ([-10\degree,-7\degree), [-7\degree,-4\degree), [-4\degree,-1\degree), (1\degree,4\degree], (4\degree,7\degree], and (7\degree,10\degree]).
We define the maximum tilted angle as 10\degree, because people prefer to take pictures without tilt unconsciously. Even if there is a tilt, it would not be very large. Besides, He $et\ al.$'s rotation fails frequently (loses image content or produces large distortions) when the rotation angle is larger than 10\degree, which affects the dataset generation.

3) Then, we filter out the rotated images with noticeable distortions manually for 3 epochs. Only 6,202 images remain from over 20$k$ samples.

4) To further enhance the quality of our dataset, we develop a mesh-based program to manually fine-tune the rotated results following He $et\ al.$'s rotation \cite{he2013content}. It allows the user to drag the mesh vertexes with a mouse to modify the mesh deformation interactively as shown in Fig. \ref{dataset_generation}{\color{red}c}. This process requires huge manual labor, and we randomly select about 20\% images from the last step to conduct the manual correction.

\begin{figure*}[!t]
    \centering
    \includegraphics[width=0.97\textwidth]{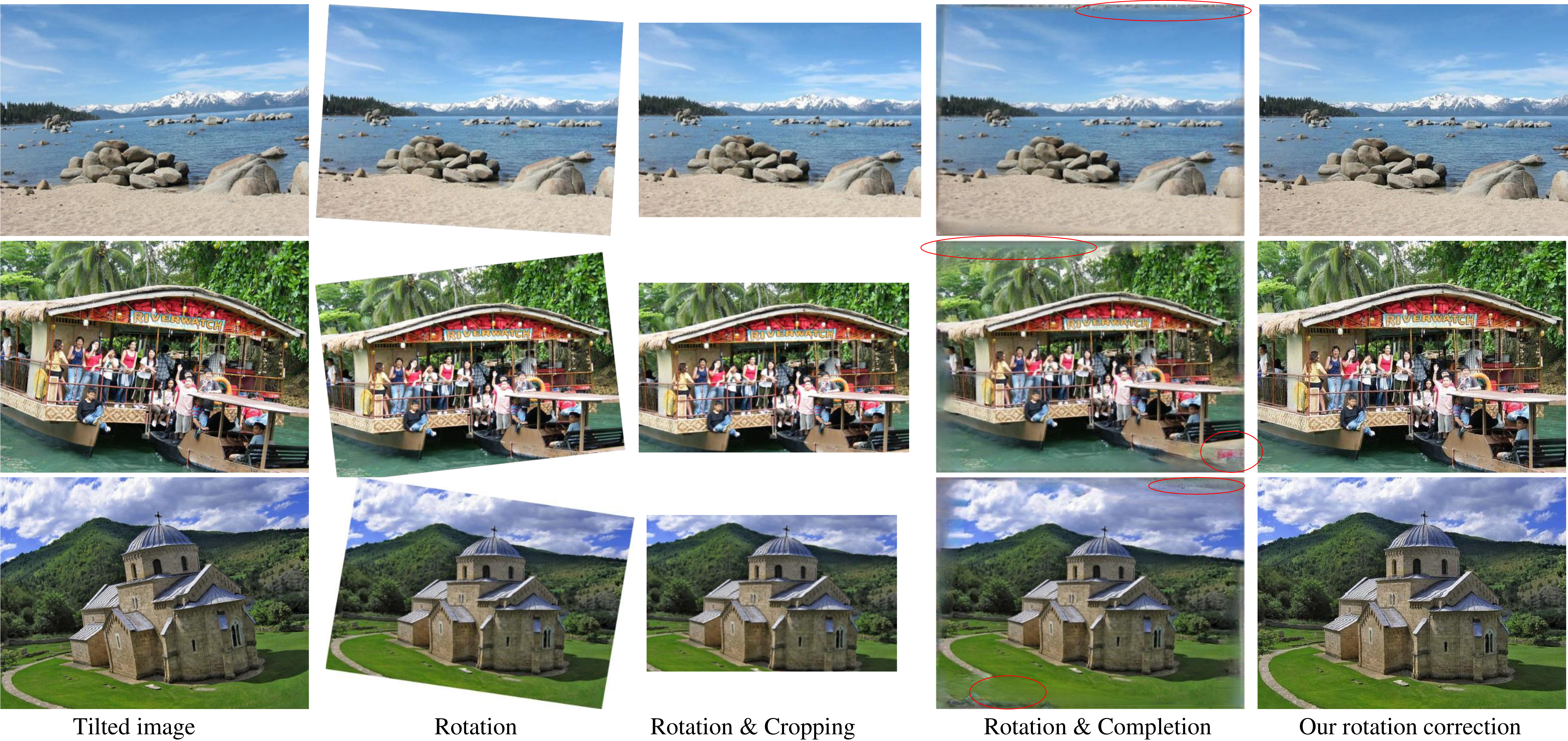}
    \vspace{-0.2cm}
    \caption{Qualitative comparisons with cropping and completion. The rotated angles are 3.5\degree, -6.5\degree, and 9.0\degree from top to bottom. The red circles highlight the failure of completion.}
    \label{vs_cropping}
\end{figure*}

\begin{figure*}[!t]
    \centering
    \includegraphics[width=0.97\textwidth]{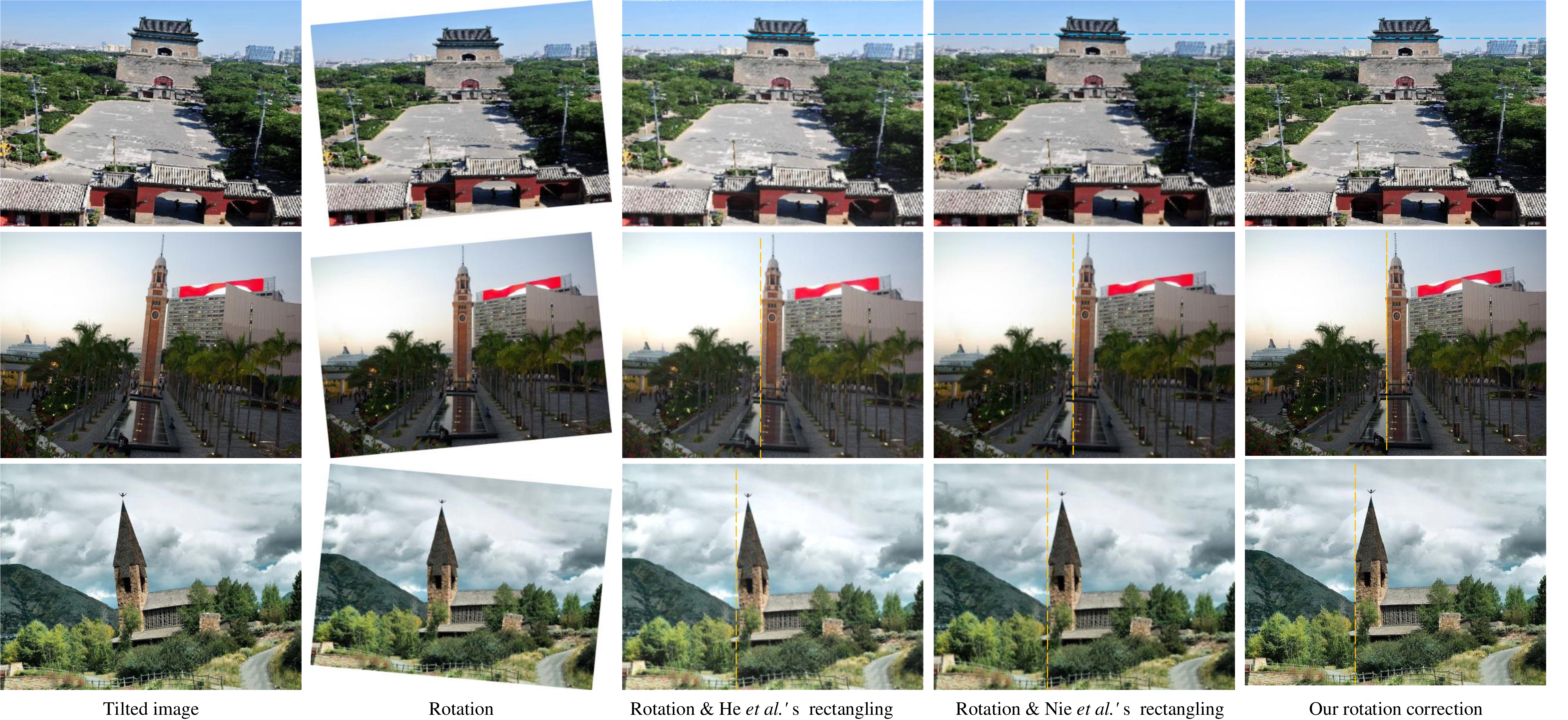}
    \vspace{-0.2cm}
    \caption{Qualitative comparisons with rectangling-based solutions. The rotated angles are -5.3\degree, -5.5\degree, and 5.4\degree from top to bottom. The rectangling algorithms take rotated images (the second col) as input and output the rectangular images at the cost of reintroducing a slight tilt in perception. We plot the horizontal or vertical dotted lines for better observation.}
    \label{vs_rectangling}
\end{figure*}

Now, we get all the samples of our dataset with tilted images, labels for corrected images, and labels for rotated angles to rectify the horizon. The training set and testing set are randomly divided according to the ratio of 9:1. Finally, we get the 5,537 samples for training and 665 samples for testing with the resolution set to $512\times 384$. We name this dataset DRC-D, and the data distribution concerning rotated angles is shown in Fig. \ref{dataset_distribution}.

\section{Experiment}
In this section, we conduct extensive comparative experiments and ablation studies in Section \ref{comparative_experiment} and \ref{ablation_study}. To further validate the effectiveness of our method, cross-dataset evaluation is demonstrated in Section \ref{comparative_experiment}. Moreover, the analysis of the feature visualization is provided to explore the secret of rotation correction in the same section. Finally, we discuss the potential applications and future prospects in Section \ref{future}.

\label{experiment}
\subsection{Implement Details}
\subsubsection{Training Details}
We use Adam optimizer \cite{kingma2014adam} to train our network with an exponentially decaying learning rate initialized to 10e-4. The batch size is set to 4 and the training process takes 150k iterations in a single GPU with NVIDIA RTX 2080 Ti.
$\omega$ and $\lambda$ are set to 0.1 and 0.25, respectively.
We assign $8\times 6$ to $U\times V$ because the high-resolution mesh regression would increase the computational burden. Moreover, the subsequent residual flow prediction can make up for the lack of pixel-wise deformation in the low-resolution mesh deformation.

\subsubsection{Inference}
The proposed algorithm can process images with arbitrary resolutions. For example, given a $2048\times 1536$ tilted image, it would be first downsampled to $512\times 384$ and the optical flows to rectify the horizon are predicted in the downsampled resolution. Then the flows would be upsampled by increasing the resolution and magnifying the values, and the corrected result can be obtained by warping the full resolution input using the upsampled flows.

It takes about 0.2 seconds to process such a high-resolution image in GPU. The running time dominantly depends on warping (interpolating) the full resolution image.

\subsection{Comparative Experiment}
\label{comparative_experiment}
\subsubsection{Compared with Content-Altering Solutions}
Cropping and completion are straightforward solutions for the irregular boundaries caused by the rigid rotation. However, cropping decreases image contents so that the cropped results would exhibit a visual effect of FoV shrinking. Completion \cite{yi2020contextual} increasing the extra contents that are visually reasonable but not reliable. Therefore, it's unfair to compare our solution with them quantitatively.

We demonstrate the qualitative comparisons in Fig. \ref{vs_cropping}, where three instances with different tilted angles are given. As the increase of rotated angles (from top to bottom), the content loss and content addition become more noticeable. Compared with them, our solution can correct the tilt naturally without content altering and angle prior.

\subsubsection{Compared with Content-Preserving Solutions}

\begin{table*}[!t]
    \centering
    \caption{Quantitative comparisons with content-preserving solutions on DIR-D.}
    \vspace{-0.1cm}
    \label{quan}
    \scalebox{0.84}{
    \begin{tabular}{llllllll}
     \toprule
     &\makecell[c]{Solution} & Angle prior& PSNR ($\uparrow$) &SSIM ($\uparrow$) &FID \cite{heusel2017gans} ($\downarrow$)&LPIPS-vgg \cite{zhang2018unreasonable} ($\downarrow$)&LPIPS-alex \cite{zhang2018unreasonable} ($\downarrow$)\\
    \cline{2-8}
     1&\makecell[c]{Rotation} & \makecell[c]{w/}& \makecell[c]{11.57} &\makecell[c]{0.374}&\makecell[c]{34.40}&\makecell[c]{0.468}&\makecell[c]{0.441}\\
     2&\makecell[c]{Rotation \& He $et\ al.$'s rectangling \cite{he2013rectangling}} & \makecell[c]{w/}& \makecell[c]{17.63} &\makecell[c]{0.488}&\makecell[c]{15.30}&\makecell[c]{0.345}&\makecell[c]{0.324}\\
     3&\makecell[c]{Rotation \& Nie $et\ al.$'s rectangling \cite{nie2022deep}} & \makecell[c]{w/}& \makecell[c]{19.89} &\makecell[c]{0.550}&\makecell[c]{13.40}&\makecell[c]{0.286}&\makecell[c]{0.295}\\
     4&\makecell[c]{He $et\ al.$'s rotation \cite{he2013content}} & \makecell[c]{w/}& \makecell[c]{\textbf{21.69}} &\makecell[c]{\textbf{0.646}}&\makecell[c]{8.51}&\makecell[c]{0.212}&\makecell[c]{0.171}\\
     5&\makecell[c]{Our rotation correction} & \makecell[c]{w/o}& \makecell[c]{21.02} &\makecell[c]{0.628}&\makecell[c]{\textbf{7.12}}&\makecell[c]{\textbf{0.205}}&\makecell[c]{\textbf{0.096}}\\

       \bottomrule
     \end{tabular}
    }
    \vspace{-0.2cm}
     \end{table*}

\begin{table*}[!t]
   \centering
   \caption{No-reference quantitative comparisons with content-preserving solutions on DIR-D.}
   \vspace{-0.1cm}
   \label{quan_noref}
   \scalebox{0.86}{
   \begin{tabular}{llllll}
    \toprule
    &\makecell[c]{Solution} & Angle prior& BRISQUE \cite{mittal2012no} ($\downarrow$) &PIQUE \cite{venkatanath2015blind} ($\downarrow$) &RankIQA \cite{liu2017rankiqa} ($\downarrow$)\\
   \cline{2-6}
    1&\makecell[c]{Rotation \& He $et\ al.$'s rectangling \cite{he2013rectangling}} & \makecell[c]{w/}& \makecell[c]{32.81} &\makecell[c]{17.20}&\makecell[c]{0.745}\\
    2&\makecell[c]{Rotation \& Nie $et\ al.$'s rectangling \cite{nie2022deep}} & \makecell[c]{w/}& \makecell[c]{32.94} &\makecell[c]{15.41}&\makecell[c]{0.795}\\
    3&\makecell[c]{He $et\ al.$'s rotation \cite{he2013content}} & \makecell[c]{w/}& \makecell[c]{28.44} &\makecell[c]{9.44}&\makecell[c]{0.
    424}\\
    4&\makecell[c]{Our rotation correction} & \makecell[c]{w/o}& \makecell[c]{\textbf{26.38}} &\makecell[c]{\textbf{8.19}}&\makecell[c]{\textbf{0.299}}\\
      \bottomrule
    \end{tabular}
   }
   \vspace{-0.2cm}
\end{table*}


\begin{figure*}[!t]
    \centering
    \includegraphics[width=0.97\textwidth]{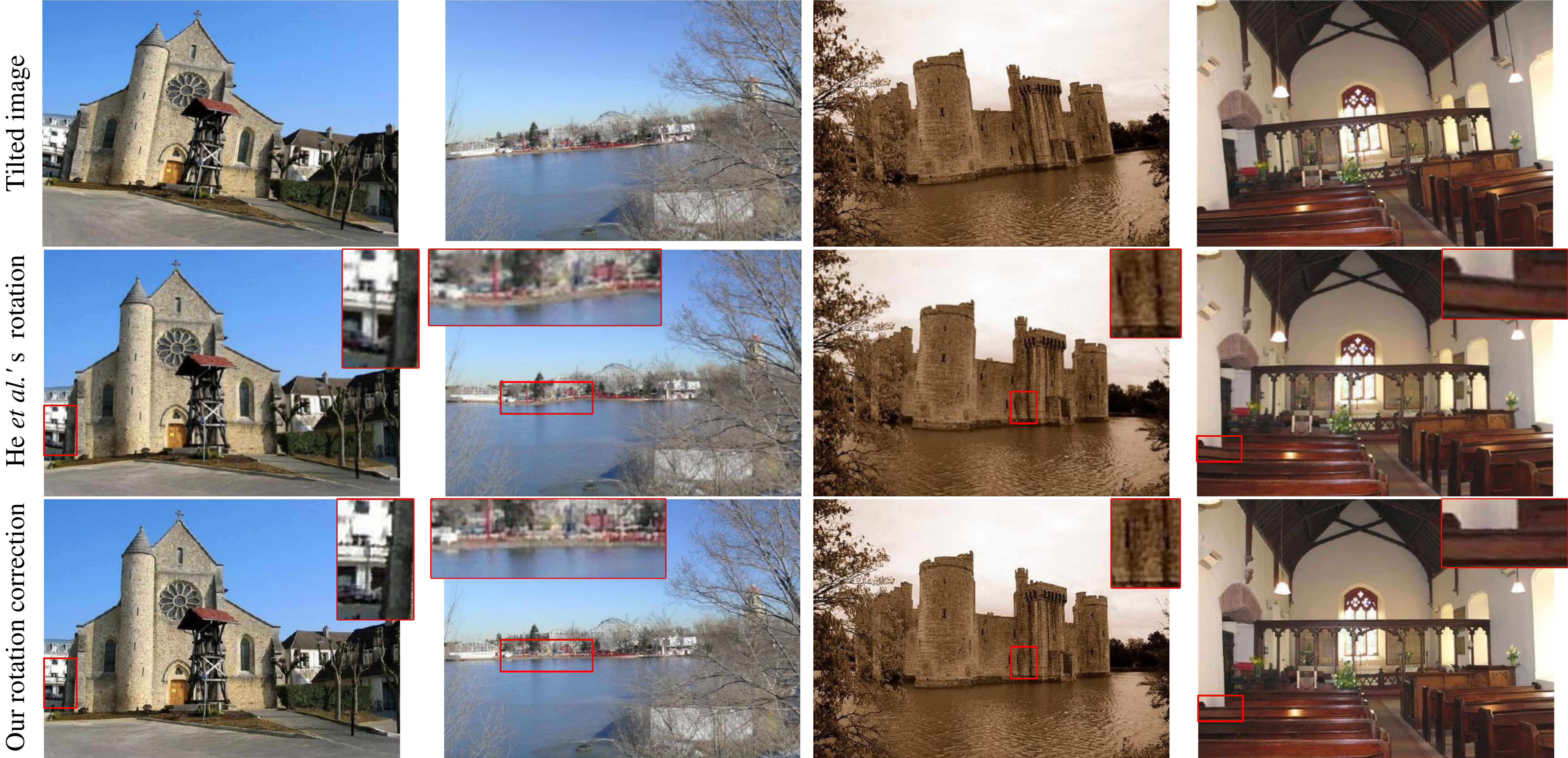}
    \vspace{-0.2cm}
    \caption{Qualitative comparison with He $et\ al.$'s rotation \cite{he2013content}. The rotated angles are -9.5\degree, 6.5\degree, 8.1\degree, and 5.7\degree from left to right. We zoomed in on local regions to compare the details. Notice that our algorithm does not require the angle prior while He $et\ al.$'s rotation \cite{he2013content} must take a specific angle as additional input. }
    \label{vs_rotation}
\end{figure*}

We also compare our solution with the content-preserving solutions as follows:

\textbf{Rotation:} The rotation operation takes the tilted image and the ground truth rotated angle as the input. After rigid rotation, the boundaries become irregular and the resolution is also changed according to the rotated angle. We resize the rotated image to the original resolution for comparison purposes.

\textbf{Rectangling:} The rectangling operation takes the rotated image as input and outputs the rectangular image without content altering. For fairness, we retrain Nie $et\ al.$'s rectangling \cite{nie2022deep} model on DRC-D by replacing the input with the rigid rotated images. We set the mesh resolution of both He $et\ al.$'s \cite{he2013rectangling} and Nie $et\ al.$'s rectangling \cite{nie2022deep} to $8\times 6$ for two reasons: 1) keeping the mesh resolution consistent with that in our solution, and 2) avoiding noticeable distortions that might be frequently produced in the condition of high-resolution mesh.

\textbf{Content-aware rotation:} Similar to rotation, the content-aware rotation algorithm \cite{he2013content} also takes the tilted image and the ground truth rotated angle as the input. But it outputs the content-rotated rectangular results. The mesh resolution is also set to $8\times 6$ since this resolution can yield better ``less-distortion'' results compared with high-resolution mesh.

We demonstrate the quantitative comparisons and no-reference quantitative comparisons in Table \ref{quan} and Table \ref{quan_noref}. From Table \ref{quan}, He $et\ al.$'s rotation is evaluated with the best PSNR and SSIM while ours is ranked the second in these two metrics.
He $et\ al.$’s rotation has an inevitable advantage on DCR-D dataset, because the tilted images are generated from horizontal images with He $et\ al$’s rotation and further manual correction. Besides, the accurate rotation angle (ground truth angle) is provided in advance for He $et\ al$’s solution, while ours requires no angle prior.
Since human eyes are sensitive to the salient regions in horizontal perception, these two statistic-based metrics cannot objectively reflect the quality of rotation correction. Therefore, we add FID \cite{heusel2017gans} and LPIPS \cite{zhang2018unreasonable} as the perceptual measures that are popularly adopted in image generation tasks \cite{yi2020contextual, esser2021taming}. From these perception-based metrics, the proposed solution reaches the best performance. Besides that, we conduct a no-reference blind image quality evaluation on these solutions. As shown in Table \ref{quan_noref}, BRISQUE \cite{mittal2012no} is a natural scene statistic-based assessment while PIQUE \cite{venkatanath2015blind} and RandIQA \cite{liu2017rankiqa} are perception-based assessments. Our method is ranked the best solution in these no-reference metrics.

\begin{figure*}[!t]
    \centering
    \includegraphics[width=0.97\textwidth]{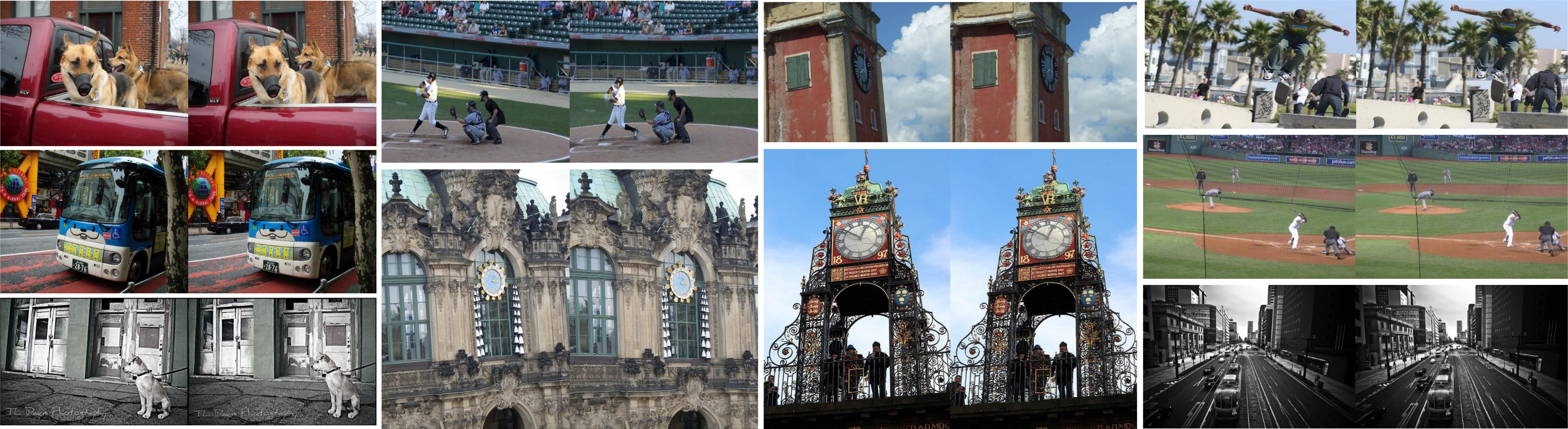}
    \vspace{-0.2cm}
    \caption{Cross-dataset evaluation. We train our model on DRC-D and qualitatively evaluate it on MS-COCO \cite{lin2014microsoft}. Several examples with different resolutions and aspect ratio are demonstrated here, where each example includes a pair of tilted image (left) and our result (right).}
    \label{cross_coco}
\end{figure*}

Moreover, the qualitative comparisons are exhibited in Fig. \ref{vs_rectangling} and Fig. \ref{vs_rotation}. The rectangling-based solutions might reintroduce the slight tilt to the rectangular results because they neglect the angle-preserving constraint. The content-aware rotation gives poor details on the rotated regions occasionally due to the limitation of mesh resolution. (Actually, higher mesh resolution in the content-aware rotation might produce much more distortions, yielding a worse performance.) On the other hand, they all require the rotated angles as additional input. Compared with them, our solution corrects the tilt naturally without this angle prior.

\subsubsection{Cross-Dataset Evaluation}

In this cross-dataset evaluation, we adopt the DRC-D dataset to train our model and test this model in other datasets. Here we conduct the testing experiments on the images from MS-COCO \cite{lin2014microsoft} and demonstrate the visual appearance. As shown in Fig. \ref{cross_coco}, several examples with different resolutions and aspect ratios are given. Our solution predicts the optical flows to eliminate the visual tilt without the angle prior.

\subsubsection{The Secret of Rotation Correction}
\begin{figure}[!t]
    \centering
    \includegraphics[width=0.48\textwidth]{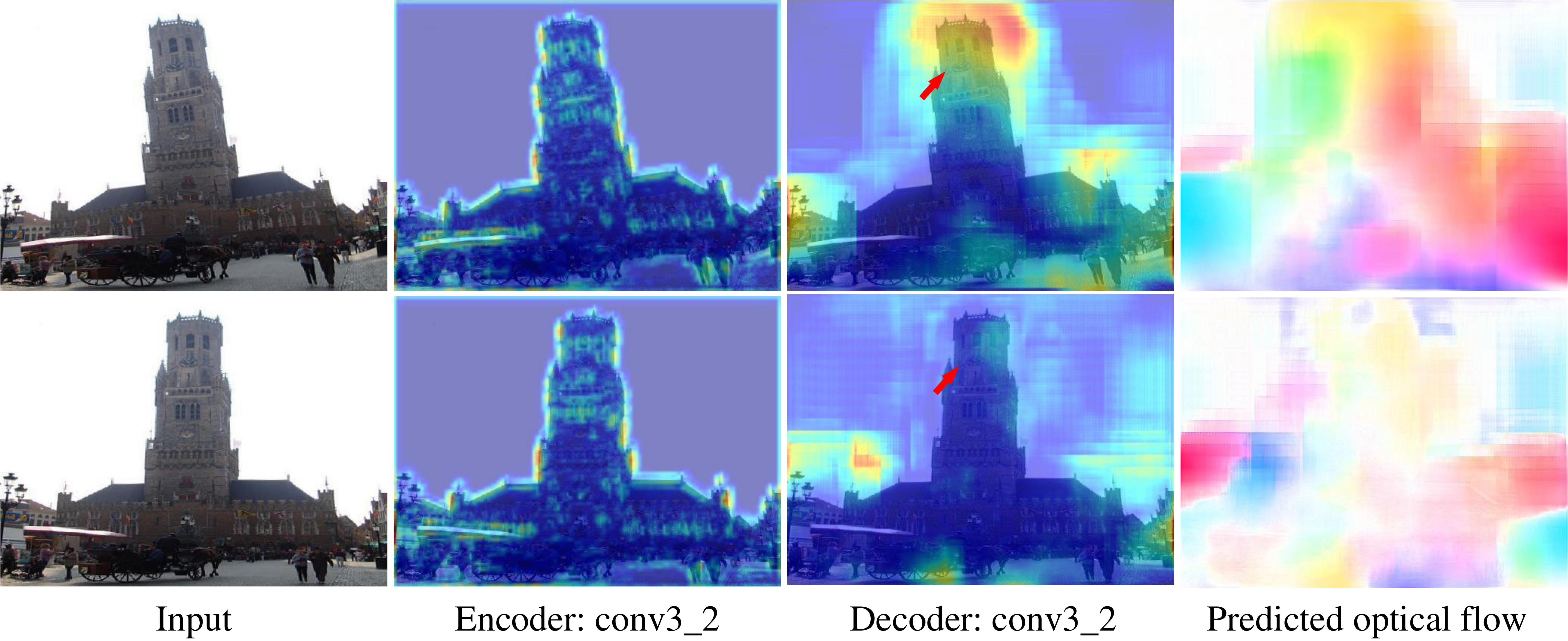}
    \vspace{-0.2cm}
    \caption{The secret of rotation correction. The endoer and decoder are responsible for different works.}
    \label{feature}
\end{figure}

\begin{figure}[!t]
    \centering
    \includegraphics[width=0.48\textwidth]{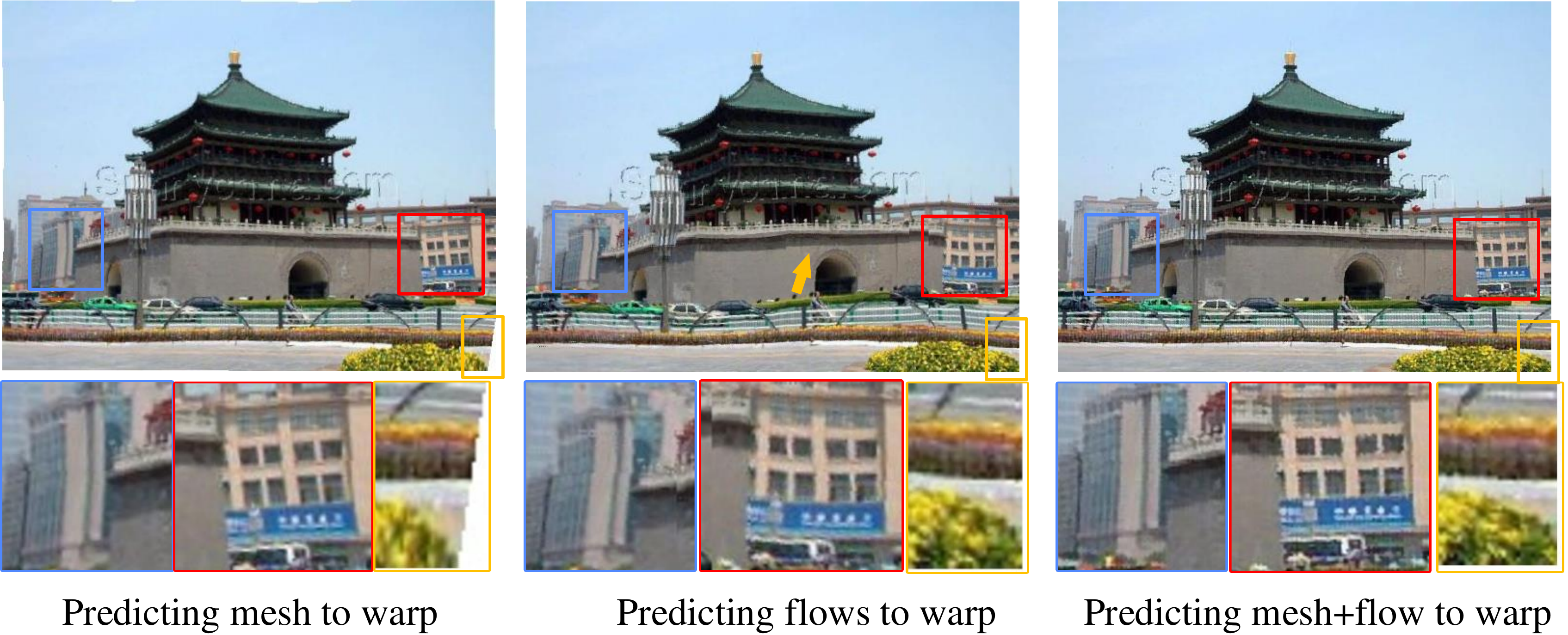}
    \vspace{-0.1cm}
    \caption{Ablation studies on warping strategies. Compared with only $mesh$/$flow$, $mesh+flow$ can yield robust correction with better details and boundaries.}
    \label{ablation_fig}
\end{figure}

To further explore how the neural network can work for rotation correction, we give an analysis of the monocular optical flow prediction process in this section. To this end, we visualize the feature maps to explore the secret of rotation correction. Specifically, we adopt an encoder-decoder network with skip connections and 5 different resolution hierarchies to predict the optical flows. The feature maps from the $conv3\_2$ layers (both encoder and decoder) and the predicted flows are visualized in Fig. \ref{feature}. The two testing samples are tilted to different degrees. By comparing the feature maps, we conclude the encoder and decoder play different roles:

1) The role of the encoder is to extract semantic features that are beneficial to horizontal perception, such as straight lines close to vertical/horizontal direction.

2) The decoder is responsible for distinguishing the tilted regions. For example, the heads of the tower in the third column of Fig. \ref{feature} are highlighted to different degrees. In the large-angle rotated instance, the network assigns larger values to the feature region (row 1, col 3, highlighted by the arrow) to predict longer flows that can correct it. In contrast, the feature region is relatively neglected in the small-rotated instance (row 2, col 3, highlighted by the arrow).

\subsection{Ablation Study}
\label{ablation_study}

\begin{table}[!t]
    \centering
    \caption{Ablation studies on DRC-D.}
    \vspace{-0.3cm}
    \label{abaltion_table}
    \scalebox{0.86}{
    \begin{tabular}{lllll}
     \toprule
     &\makecell[c]{Architecture} &\makecell[c]{Loss} & PSNR ($\uparrow$) & SSIM ($\uparrow$) \\
    \cline{2-5}
     1&\makecell[c]{Mesh} &\makecell[c]{$\mathcal{L}_{content}$}& \makecell[c]{19.64} &\makecell[c]{0.618}\\
     2&\makecell[c]{Mesh+mesh} &\makecell[c]{$\mathcal{L}_{content}$}& \makecell[c]{19.52} &\makecell[c]{0.604}\\
     3&\makecell[c]{Flow} &\makecell[c]{$\mathcal{L}_{content}$}& \makecell[c]{18.48} &\makecell[c]{0.500}\\
     4&\makecell[c]{Flow+flow} &\makecell[c]{$\mathcal{L}_{content}$}& \makecell[c]{18.66} &\makecell[c]{0.511}\\
     5&\makecell[c]{Mesh+flow} &\makecell[c]{$\mathcal{L}_{content}$}& \makecell[c]{20.90} &\makecell[c]{0.621}\\
     6&\makecell[c]{Mesh+flow} &\makecell[c]{$\mathcal{L}_{content}+\mathcal{L}_{symmetry}$}& \makecell[c]{\textbf{21.02}} &\makecell[c]{\textbf{0.628}}\\
       \bottomrule
     \end{tabular}
    }
    \vspace{-0.2cm}
 \end{table}

The proposed monocular optical flow prediction strategy is simple but effective. We evaluate the effectiveness of every module on DRC-D.

\subsubsection{Only Mesh}
Another effective approach to correct the tilt is to predict the mesh deformation instead of the optical flows. However, the low-resolution mesh deformation might damage the content details while the high-resolution one tends to cause unnatural
distortions to the mesh, such as self-intersection. Besides, only predicting the mesh can yield uneven rectangular boundaries in the corrected image as shown in Fig. \ref{ablation_fig} (left).
Considering the fairness of comparing the mesh prediction method with ours, we also repeat the mesh deformation module to replace the residual optical flow prediction module (``mesh+mesh'' in Table \ref{abaltion_table}).

\subsubsection{Only Flow}
Predicting pixel-wise optical flows can produce a pixel-wise warp. But this warp is extremely unstable especially in large-angle tilted scenes as shown in Fig. \ref{ablation_fig} (middle). On the other hand, it can produce perfect rectangular boundaries in the corrected result, which is opposite to the mesh warp. Also, we repeat the optical flow prediction module to replace the mesh deformation estimation module (``flow+flow'' in Table \ref{abaltion_table}).

\subsubsection{Mesh + Flow}
Mesh prediction can produce a robust warp, but it lacks pixel-wise deformation capability and produces uneven boundaries. In contrast, flow prediction can produce pixel-wise flexible warp with even rectangular boundaries, but the prediction would become unstable in large-angle tilted scenes. We combine their advantages to form a robust elastic warp that can rectify the horizon effectively.

\subsubsection{Symmetry-Equivariant Loss}
The symmetry-equivariant loss can further improve the performance to rectify the horizon. Besides, it can enhance the generalization capability when the pre-trained model is transferred to other datasets.


\subsection{Future Prospect}
\label{future}
To correct the tilt, existing solutions share a two-stage pipeline, in which a single image calibration method is first used to estimate the tilt angle and then a content-aware warp method is used to remove the content tilt. Compared with them, we propose the first one-stage baseline by constructing a robust elastic warp and benchmark dataset. For future works, more geometric features (e.g., line, curve, et al.) can be combined with the semantic features to reach better content preservation. To generalize to other scenes, weakly-supervised and semi-supervised algorithms could be studied to decrease the urgent demand for the expensive labeled data. Besides, the proposed mesh-to-flow strategy shows the potential to be extended to other image-warping tasks, such as portrait correction\cite{tan2021practical}, image rectangling\cite{nie2022deep}, image retargeting\cite{tan2019cycle}, and so on.

\section{Conclusion}
\label{conclusion}
In this paper, we propose a new and practical task to automatically correct the content tilt without the angle prior, which can be easily integrated into image editing applications. To accomplish this task, we design a simple but effective warping strategy. Particularly, we combine the robustness of mesh estimation and the flexibility of optical flow prediction into a unified framework, contributing to a robust elastic warp. To establish an evaluation benchmark and train the learning framework, we build a comprehensive rotation correction dataset with a large diversity in rotated angles and scenes. Finally, we validate our method by conducting extensive experiments and ablation studies. The results show our superiority over other state-of-the-art solutions and the effectiveness of the new warping strategy.

\normalem
\bibliographystyle{ieeetr}
\bibliography{reference}

\begin{thebibliography}{10}

\bibitem{yi2020contextual}
Z.~Yi, Q.~Tang, S.~Azizi, D.~Jang, and Z.~Xu, ``Contextual residual aggregation
  for ultra high-resolution image inpainting,'' in {\em Proceedings of the
  IEEE/CVF Conference on Computer Vision and Pattern Recognition},
  pp.~7508--7517, 2020.

\bibitem{he2013rectangling}
K.~He, H.~Chang, and J.~Sun, ``Rectangling panoramic images via warping,'' {\em
  ACM Transactions on Graphics}, vol.~32, no.~4, pp.~1--10, 2013.

\bibitem{li2015geodesic}
D.~Li, K.~He, J.~Sun, and K.~Zhou, ``A geodesic-preserving method for image
  warping,'' in {\em Proceedings of the IEEE/CVF Conference on Computer Vision
  and Pattern Recognition}, pp.~213--221, 2015.

\bibitem{zhang2020content}
Y.~Zhang, Y.-K. Lai, and F.-L. Zhang, ``Content-preserving image stitching with
  piecewise rectangular boundary constraints,'' {\em IEEE Transactions on
  Visualization and Computer Graphics}, vol.~27, no.~7, pp.~3198--3212, 2020.

\bibitem{nie2022deep}
L.~Nie, C.~Lin, K.~Liao, S.~Liu, and Y.~Zhao, ``Deep rectangling for image
  stitching: A learning baseline,'' in {\em Proceedings of the IEEE/CVF
  Conference on Computer Vision and Pattern Recognition}, pp.~5740--5748, 2022.

\bibitem{wu2022rectangling}
J.-L. Wu, J.-J. Shi, and L.~Zhang, ``Rectangling irregular videos by optimal
  spatio-temporal warping,'' {\em Computational Visual Media}, vol.~8, no.~1,
  pp.~93--103, 2022.

\bibitem{he2013content}
K.~He, H.~Chang, and J.~Sun, ``Content-aware rotation,'' in {\em Proceedings of
  the IEEE/CVF International Conference on Computer Vision}, pp.~553--560,
  2013.

\bibitem{von2008lsd}
R.~G. Von~Gioi, J.~Jakubowicz, J.-M. Morel, and G.~Randall, ``Lsd: A fast line
  segment detector with a false detection control,'' {\em IEEE Transactions on
  Pattern Analysis and Machine Intelligence}, vol.~32, no.~4, pp.~722--732,
  2008.

\bibitem{9605632}
L.~Nie, C.~Lin, K.~Liao, S.~Liu, and Y.~Zhao, ``Depth-aware multi-grid deep
  homography estimation with contextual correlation,'' {\em IEEE Transactions
  on Circuits and Systems for Video Technology}, pp.~1--1, 2021.

\bibitem{gallagher2005using}
A.~C. Gallagher, ``Using vanishing points to correct camera rotation in
  images,'' in {\em The 2nd Canadian Conference on Computer and Robot Vision},
  pp.~460--467, 2005.

\bibitem{lee2021ctrl}
J.~Lee, H.~Go, H.~Lee, S.~Cho, M.~Sung, and J.~Kim, ``Ctrl-c: Camera
  calibration transformer with line-classification,'' in {\em Proceedings of
  the IEEE/CVF International Conference on Computer Vision}, pp.~16228--16237,
  2021.

\bibitem{lin2022deep}
Y.~Lin, R.~Wiersma, S.~L. Pintea, K.~Hildebrandt, E.~Eisemann, and J.~C. van
  Gemert, ``Deep vanishing point detection: Geometric priors make dataset
  variations vanish,'' {\em arXiv preprint arXiv:2203.08586}, 2022.

\bibitem{xian2019uprightnet}
W.~Xian, Z.~Li, M.~Fisher, J.~Eisenmann, E.~Shechtman, and N.~Snavely,
  ``Uprightnet: geometry-aware camera orientation estimation from single
  images,'' in {\em Proceedings of the IEEE/CVF International Conference on
  Computer Vision}, pp.~9974--9983, 2019.

\bibitem{lee2013automatic}
H.~Lee, E.~Shechtman, J.~Wang, and S.~Lee, ``Automatic upright adjustment of
  photographs with robust camera calibration,'' {\em IEEE Transactions on
  Pattern Analysis and Machine Intelligence}, vol.~36, no.~5, pp.~833--844,
  2013.

\bibitem{do2020surface}
T.~Do, K.~Vuong, S.~I. Roumeliotis, and H.~S. Park, ``Surface normal estimation
  of tilted images via spatial rectifier,'' in {\em European Conference on
  Computer Vision}, pp.~265--280, 2020.

\bibitem{tan2021practical}
J.~Tan, S.~Zhao, P.~Xiong, J.~Liu, H.~Fan, and S.~Liu, ``Practical wide-angle
  portraits correction with deep structured models,'' in {\em Proceedings of
  the IEEE/CVF Conference on Computer Vision and Pattern Recognition},
  pp.~3498--3506, 2021.

\bibitem{tan2019cycle}
W.~Tan, B.~Yan, C.~Lin, and X.~Niu, ``Cycle-ir: Deep cyclic image
  retargeting,'' {\em IEEE Transactions on Multimedia}, vol.~22, no.~7,
  pp.~1730--1743, 2019.

\bibitem{kweon2021pixel}
H.~Kweon, H.~Kim, Y.~Kang, Y.~Yoon, W.~Jeong, and K.-J. Yoon, ``Pixel-wise deep
  image stitching,'' {\em arXiv preprint arXiv:2112.06171}, 2021.

\bibitem{ronneberger2015u}
O.~Ronneberger, P.~Fischer, and T.~Brox, ``U-net: Convolutional networks for
  biomedical image segmentation,'' in {\em International Conference on Medical
  Image Computing and Computer-Assisted Intervention}, pp.~234--241, 2015.

\bibitem{hartley2003multiple}
R.~Hartley and A.~Zisserman, {\em Multiple view geometry in computer vision}.
\newblock Cambridge university press, 2003.

\bibitem{johnson2016perceptual}
J.~Johnson, A.~Alahi, and L.~Fei-Fei, ``Perceptual losses for real-time style
  transfer and super-resolution,'' in {\em European Conference on Computer
  Vision}, pp.~694--711, 2016.

\bibitem{simonyan2014very}
K.~Simonyan and A.~Zisserman, ``Very deep convolutional networks for
  large-scale image recognition,'' {\em arXiv preprint arXiv:1409.1556}, 2014.

\bibitem{deng2009imagenet}
J.~Deng, W.~Dong, R.~Socher, L.-J. Li, K.~Li, and L.~Fei-Fei, ``Imagenet: A
  large-scale hierarchical image database,'' in {\em Proceedings of the
  IEEE/CVF Conference on Computer Vision and Pattern Recognition},
  pp.~248--255, 2009.

\bibitem{kingma2014adam}
D.~P. Kingma and J.~Ba, ``Adam: A method for stochastic optimization,'' {\em
  arXiv preprint arXiv:1412.6980}, 2014.

\bibitem{heusel2017gans}
M.~Heusel, H.~Ramsauer, T.~Unterthiner, B.~Nessler, and S.~Hochreiter, ``Gans
  trained by a two time-scale update rule converge to a local nash
  equilibrium,'' {\em Advances in Neural Information Processing Systems},
  vol.~30, 2017.

\bibitem{zhang2018unreasonable}
R.~Zhang, P.~Isola, A.~A. Efros, E.~Shechtman, and O.~Wang, ``The unreasonable
  effectiveness of deep features as a perceptual metric,'' in {\em Proceedings
  of the IEEE/CVF conference on computer vision and pattern recognition},
  pp.~586--595, 2018.

\bibitem{mittal2012no}
A.~Mittal, A.~K. Moorthy, and A.~C. Bovik, ``No-reference image quality
  assessment in the spatial domain,'' {\em IEEE Transactions on Image
  Processing}, vol.~21, no.~12, pp.~4695--4708, 2012.

\bibitem{venkatanath2015blind}
N.~Venkatanath, D.~Praneeth, M.~C. Bh, S.~S. Channappayya, and S.~S. Medasani,
  ``Blind image quality evaluation using perception based features,'' in {\em
  2015 Twenty First National Conference on Communications}, pp.~1--6, 2015.

\bibitem{liu2017rankiqa}
X.~Liu, J.~Van De~Weijer, and A.~D. Bagdanov, ``Rankiqa: Learning from rankings
  for no-reference image quality assessment,'' in {\em Proceedings of the
  IEEE/CVF International Conference on Computer Vision}, pp.~1040--1049, 2017.

\bibitem{esser2021taming}
P.~Esser, R.~Rombach, and B.~Ommer, ``Taming transformers for high-resolution
  image synthesis,'' in {\em Proceedings of the IEEE/CVF Conference on Computer
  Vision and Pattern Recognition}, pp.~12873--12883, 2021.

\bibitem{lin2014microsoft}
T.-Y. Lin, M.~Maire, S.~Belongie, J.~Hays, P.~Perona, D.~Ramanan,
  P.~Doll{\'a}r, and C.~L. Zitnick, ``Microsoft coco: Common objects in
  context,'' in {\em European Conference on Computer Vision}, pp.~740--755,
  2014.

\end{thebibliography}

\end{document}